\documentclass{bmvc2k}
\usepackage{mathtools}
\usepackage{amsmath}
\usepackage{bbm}

\title{Training face verification models from generated face identity data}

\addauthor{Dennis Conway}{dennis.john.conway@gmail.com}{1}
\addauthor{Loic Simon}{loic.simon@ensicaen.fr}{1}
\addauthor{Alexis Lechervy}{alexis.lechervy@unicaen.fr}{1}
\addauthor{Frederic Jurie}{frederic.jurie@safrangroup.com}{2}

\addinstitution{
 Normandie Univ, UNICAEN, ENSICAEN, CNRS UMR GREYC\\
 6 Boulevard Maréchal Juin, CS 45053, 14050 CAEN cedex 4, France
}
\addinstitution{
 Safran\\
 Rue des jeunes Bois, 78117 Châteaufort
}
\runninghead{Conway, Jurie, Simon, Lechervy}{ Training from generated data}

\begin{document}

\maketitle

\begin{abstract}
Machine learning tools are becoming increasingly powerful and widely used. Unfortunately membership attacks, which seek to uncover information from data sets used in machine learning, have the potential to limit data sharing. In this paper we consider an approach to increase the privacy protection of data sets, as applied to face recognition. Using an auxiliary face recognition model, we build on the StyleGAN generative adversarial network and feed it with latent codes combining two distinct sub-codes, one encoding visual identity factors, and, the other, non-identity factors. By independently varying these vectors during image generation, we create a synthetic data set of fictitious face identities. We use this data set to train a face recognition model. The model performance degrades in comparison to the state-of-the-art of face verification. When tested with a simple membership attack our model provides good privacy protection, however the model performance degrades in comparison to the state-of-the-art of face verification. 
We find that the addition of a small amount of private data greatly improves the performance of our model, which highlights the limitations of using synthetic data to train machine learning models.
\end{abstract}

\section{Introduction}
In recent times, the availability and power of large data sets has increased rapidly in a variety of different communities. With this influx of data come new challenges. Many data sets include sensitive information, in particular medical, biometric and financial data. It is insufficient to de-identify these data sets, as it has been shown that it is possible to re-identify entries in the data set using comparison with third-party data sets \citep{henriksen2016re}. One solution to these privacy problems in the application of training machine learning algorithms is to create a sanitised synthetic data set in place of a real one. As there is no requirement that a data set consists of real samples, the synthetic data set merely needs to have the same statistical properties as its source. 

There are then two matters of interest: firstly, how well does the synthetic data set approximate the real data set that it replaces, and secondly, how well does the sanitised data set protect the sensitive data of the real data set. The first question can be evaluated by comparing the performance of an algorithm trained on the synthetic data set with one trained on the underlying real data set. We would like to have both methods to give similar levels of performance, as this would show that for our intended application the two data sets have similar statistical properties. 

In terms of the second question, it is difficult to ascertain the level of privacy gained by creating a synthetic data set. If a method is used which employs $\epsilon-$differential privacy, then the privacy budget, $\epsilon$, can be used as a metric for how well the data is protected, however it is not always practical to use $\epsilon-$differential privacy methods when working with image generation.
Indeed, it was noted in \cite{saxena2021generative} that generators trained in such a differentially-private way fail to maintain a trade-off between sample quality and 
diversity. It has been shown in recent work that under some assumptions, synthetic data samples generated using GANs are inherently $(\epsilon, \delta)-$ differentially private \citep{lin2021privacy}.

For our work we consider the scenario where we would like a data set of faces which is private in terms of face identity, with application to face verification. The data sets commonly used for this task are composed of faces from celebrity individuals, whose identities are supposed to be in public domain. This is not always the case however, as the large number of identities needed for training face recognition algorithms requires a loose definition of celebrity, such that many of the members of the data sets object to their inclusion \citep{prabhakar2003biometric, senior2011privacy}. Privacy concerns become even more clear when we consider private databases such as those owned by Google and Facebook \citep{naker2017now}.
To build our synthetic data set we build on the StyleGAN2 architecture \citep{karras2020analyzing}. StyleGAN2 is a powerful Generative Adversarial Network (GAN) that when trained on the FFHQ dataset \citep{karrasflickr} can produce photorealistic random faces belonging to random identities. Using StyleGAN2, we propose a method which allows us to generate multiple samples from a random identity and therefore enables the creation of a face identity dataset. With part of the StyleGAN2 latent code fixed, we generate multiple images and penalise dissimilarity in facial features using the pre-trained state-of-the-art face recognition algorithm ArcFace \citep{deng2019arcface}. We evaluate our synthetic data set on the task of training a face recognition algorithm.

Compared to models from real data sets, the model trained using our generated synthetic data set shows a drop of performance in the task of face recognition. We show however that better performance is achieved when our synthetic data set is supplemented with an existing real data set. Using a simple membership attack, we show that our model trained with synthetic data provides some degree of privacy protection relative to the real data model.

\section{Related work}

The idea of using GANs to generate synthetic datasets is not new. It has been explored, for example in \citet{choi2017generating} where synthetic Electronic Health Records are generated using a GAN. Their results show that the synthetic data is comparable to the real data in terms of distribution statistics, performance in predictive modeling tasks, and by verification by an expert review. They observe an empirical privacy gain from their method. An advancement of this idea was made by \citet{jordon2018pate} with their algorithm PATE-GAN, which quantified a privacy gain by introducing a layer of differential privacy. Their algorithm was based around a "student" GAN discriminator which was taught to label synthetic examples as "real" or "fake" based on the predictions of an array of "parent" discriminators. The parent discriminators are trained on sections of the real data. The student however is trained on synthetic data which has been labelled real or fake using a differentially private vote aggregation from the parent discriminators. The generator is trained  using only the differentially private discriminator, so by the post-processing rule of differential privacy, the generated dataset has the same privacy guarantees as the private discriminator \citep{dwork2014algorithmic}. The results work well using a variety of different datasets. The method has not be shown to work well with image datasets however, possibly due to the higher complexity of the data.

Compared to PATE-GAN and other work which involves conditional GANs such as \citet{odena2017conditional} and \citet{trigueros2018generating}, our work has the added constraint that for our synthetic dataset we would like to create not just synthetic samples but also synthetic classes (our identities). This issue has been explored by \citet{donahue2017semantically} who propose Semantically Decomposed GANs (SD-GANs) which encourage the disentanglement of the latent space of GANs. Their GAN is trained using a latent code $z$ decomposed into $z_1$ and $z_2$. Two images are then produced using a fixed $z_1$ and varying $z_2$. A discriminator is then tasked with differentiating between the pair of fake images and a pair of images from the real dataset belonging to a single identity. Using a facial verification tool, the authors show that their method is successful in producing images belonging to the same identity, however in contrast with our work there is no investigation of its performance in the task of training a face recognition algorithm.

\section{Background}

\subsection{Generative Adversarial Networks}

In a GAN a generative network $G$ takes an input $\mathbf{z} \in Z$ drawn from a known distribution $P_z$ and produces $G(z)$, which for our case is an image $I$. The generator learns to produce images which match those drawn from a target distribution $P_r$ by way of a minmax game played between the $G$ and a discriminator $D$. In the most simple formulation this takes the form of:
\begin{equation}
\min_G\max_D V(G, D) = \mathbbm{E}_{\mathbf{x}\sim P_r}[\log D(\mathbf{x})] + \mathbbm{E}_{\mathbf{z}\sim P_z}[\log (1 - D(G(\mathbf{z}))].
\label{eq:gan}
\end{equation}

\citep{karras2019style} introduced the idea of StyleGAN. In StyleGAN and its successor StyleGAN2, the generator can be considered split into a style network and a synthesis network. As with a traditional GAN, the style network takes an input $\mathbf{z}$ drawn from a distribution $P_z$. The network then maps $z$ to another higher dimensional space $W$, which are called "styles". This higher dimensional latent code is then input to the synthesis network to generate the image $I$. Due to the mixing of different styles during training and path length regularisation, the $W$-space latent codes of StyleGAN able to achieve better native disentanglement compared to a regular GAN.

\subsection{Face recognition networks}

Face recognition is a task that is predominately performed using deep convolutional neural networks to create an embedding from an input face. The adjacent task of face verification (determining if two face images belong to the same identity or not) can then be performed by assessing the distance between two embeddings. There are two main methods for training this embedding, either by directly learning an embedding, as in the case of the triplet loss, or by training a multi-class classifier. The latter is the method of choice in the algorithm ArcFace \citep{deng2019arcface}, which we use in this paper.

\section{Synthetic data set generation}
\begin{figure}
\centering
\includegraphics[width=\textwidth]{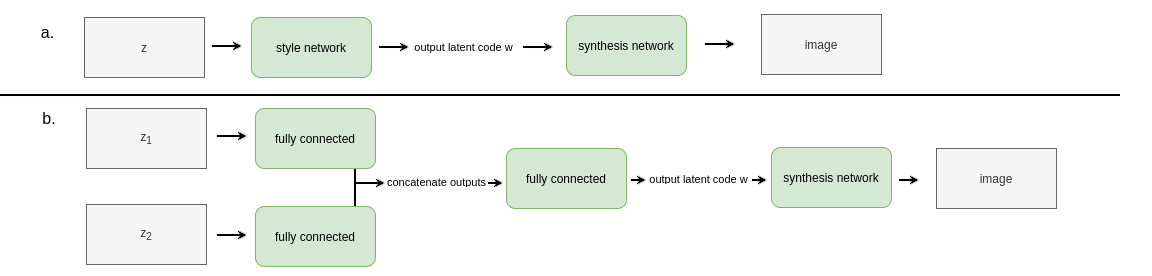}
\caption{a. the architecture of StyleGAN2 and b. the new architecture that we train to separate identity from other factors.}
\label{fig:arch1}

\end{figure}

The  model we consider is similar to the work of \citet{donahue2017semantically} in their SD-GANs, but applied to StyleGAN2. We begin with a generator $G$ that takes two input latent codes, $z_1 \in \mathbbm{R}^{512}$ and $z_2 \in \mathbbm{R}^{512}$, both drawn from a standard normal distribution. The codes $z_1$ and $z_2$ are intended to correspond to identity and non-identity features respectively. Both input codes are then passed through separate fully connected layers which are trained independently. They are then concatenated to form a layer of size $\mathbbm{R}^{1024}$ which is the reduced to an embedding layer $\in \mathbbm{R}^{512}$ from another fully connected layer. A schematic of the architecture is shown in Figure~\ref{fig:arch1}.

During training three samples are generated for each sample in the batch: an anchor with completely random factors $z_{1,0}$ and $z_{2,0}$, another realisation of the same identity $z_{1,0}$ with new other factors $z_{2,+}$ and a third image from a different identity $z_{1,-}$ with the same non-identity factors $z_{2,0}$ as from the anchor.

The GAN is trained using almost the exact same method as in the original StyleGAN2 with a size of 256x256 pixels and using the FFHQ dataset for real images. The training considers only the anchor images to calculate GAN losses. 

The loss however is augmented by a penalty which forces $z_1$ and $z_2$ into identity and non-identity factors using two auxiliary networks: a pose network $P$ which, operating on a face, returns the 3-angle pose position of the face \citep{ruiz2018fine}, and the ArcFace face embedding network $F$ \citep{deng2019arcface} which returns a face identity embedding in $\mathbbm{R}^{512}$. Designating the images as $I_0$ for anchor, $I_+$ for fixed identity and $I_-$ for different identity images, the new losses which we introduce to the existing GAN losses are:
\begin{equation}
\begin{split}
L_P^- = \lVert{P(I_0) - P(I_-)}\rVert^2, &\:
L_P^+ = -\lVert{P(I_0) - P(I_+)}\rVert^2 \\ 
L_F^- = -\arccos{(F(I_0) \cdot F(I_+))}, &\:
L_F^+ = \arccos{(F(I_0) \cdot F(I_-))}.
\end{split}
\end{equation}
These losses are added to generator loss (cf. Eq.\eqref{eq:gan}) with a weighting strength of $\lambda = 0.1$.
The different losses are clamped so that they do not effect intra-class diversity once $z_1$ and $z_2$ have been sufficiently disentangled. The clamping is as follows:
\begin{equation}
\begin{split}
    L_P^+ = \max(L_P^+, 3.0), &\:
    L_P^- = \min(L_P^-, 5.0) \\ 
    L_F^- = \max(L_F^-, 0.70), &\:
    L_F^+ = \min(L_F^+, 1.40).
\end{split}
\end{equation}


\section{Results}

\begin{figure}
\centering
\includegraphics[width=1.\textwidth]{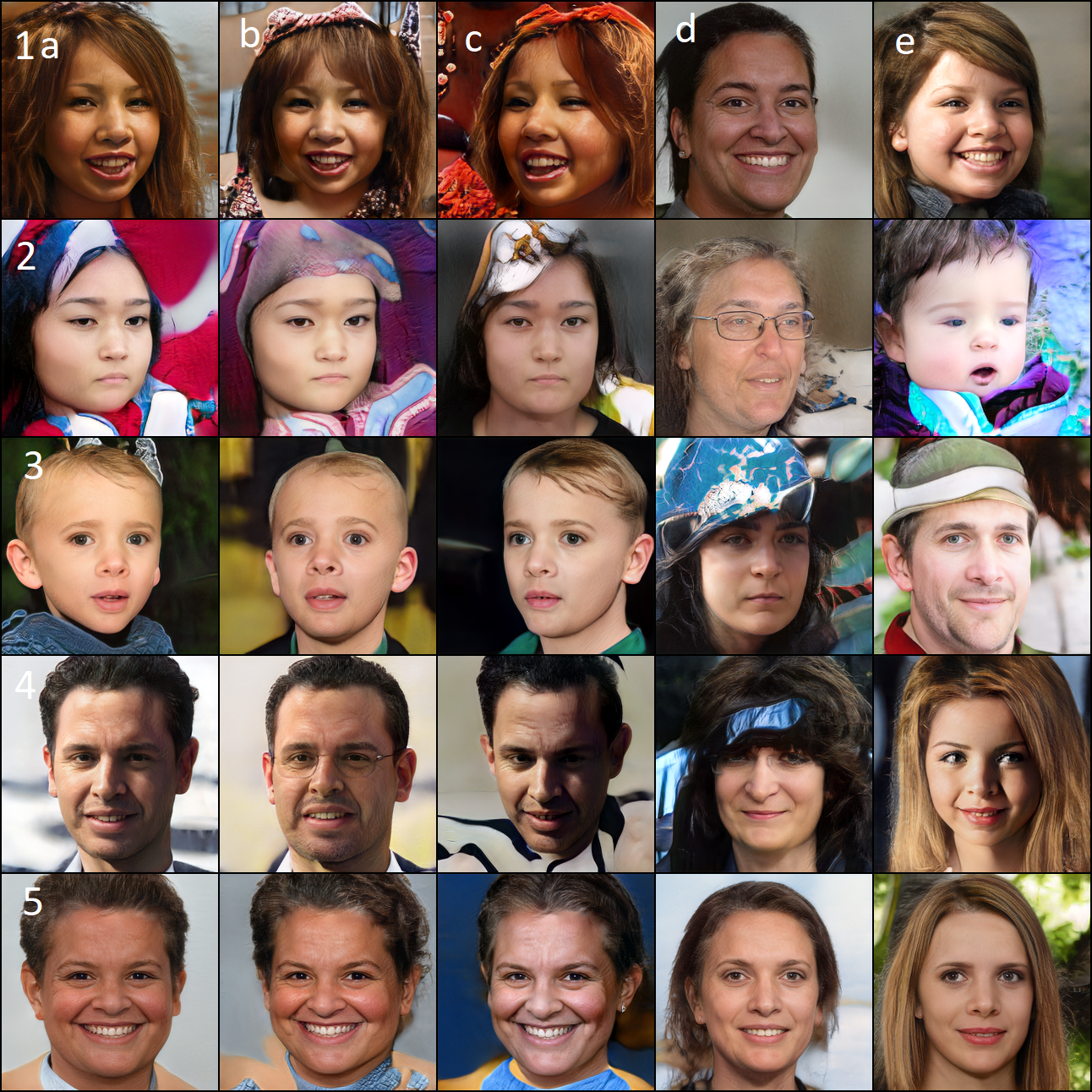}
\caption{Randomly generated realisations from our model. Column a. shows a set of five anchor images. Columns b. and c. show two realisations from varying the pose input for each anchor image with a fixed identity input. Columns c. and d. show two realisations from varying the identity input and keeping the pose input fixed.}
\label{fig:faces1}
\end{figure}

An unlimited number of hallucinations from an identity can now be generated by fixing $z_1$ and varying $z_2$. A plot showing example generations from randomly generated identities is shown in Figure~\ref{fig:faces1}.

To generate a dataset for face recognition, a set of 100 realisations are generated for each of 10,000 different identities. The GAN is trained for 150k iterations, where-after there is no gain in the quality of images generated. The dataset is then prepared for face recognition training by aligning and cropping with MTCNN \cite{Zhang16} to $112 \times 112$ pixels. In comparison the dataset MS-Celeb-1M contains 6.46 million images from 94,682 identities \citep{guo2016ms} and VGGFace2 contains 3.31 million images from 9,131 identities \citep{cao2018vggface2}.

The dataset is used to train ArcFace using the default settings: a latent space size of $\mathbbm{R}^{512}$ and an angular margin of 0.5. The training is terminated after 10 epochs without improvement on the test set of LFW. The trained model is then evaluated on LFW and achieves a performance of 83.1\% accuracy.

\subsection{Auxiliary data}
To better understand the performance of the synthetic data set, we retrain the ArcFace face recognition algorithm using the synthetic data supplemented with varying quantities of real data from the MS-Celeb-1M data set. The 10,000 synthetic identities are combined with a random selection of the 94,682 identities in MS-Celeb-1M and used to train ArcFace using the same parameters as in the previous experiment.

We find that even a small amount of real data can greatly improve the performance of our synthetic data set. We observe that the models trained using real and synthetic data outperform models trained on the same subset of real data without the synthetic data. With a sufficiently large split we see that models trained on solely the real data set begin to outperform models using synthetic and real data. At a larger split this becomes less noticeable as the size of the real data set begins to dominate the synthetic data set.

\begin{figure}
\centering
\includegraphics[width=0.6\textwidth]{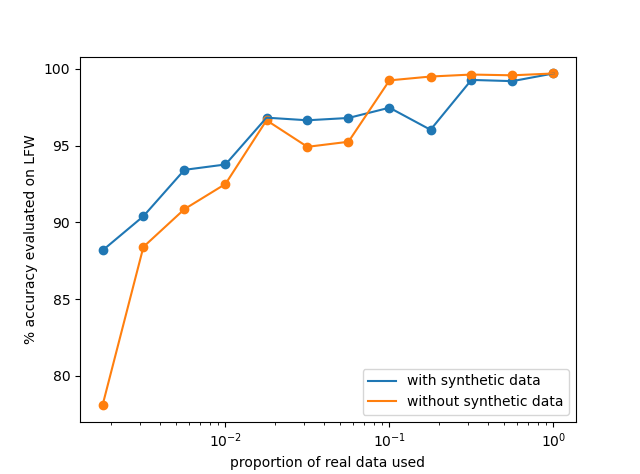}
\caption{Performance of face recognition models trained on varying splits of real and synthetic data.}
\label{fig:aux}
\end{figure}

\subsection{Membership attack}

We carry out a simple membership attack to show the improved privacy from our synthetic data set. We assume that we are given black-box access to 1) the ArcFace model trained on MS-Celeb-1M which was used in training our modified StyleGAN2 2) the ArcFace model trained on our synthetic data. Given an identity, the goal of a membership attack is to determine if the identity was used in the training of the model. Our membership attack seeks to exploit the difference in prediction entropy of a target model between its training set and a test set, as per \cite{shokri2017membership}.

We take one sample from each of the 5749 identities in the LFW data set and an equal number of distinct identities randomly from MS-Celeb-1M. Both sets of images are evaluated using ArcFace to produce a feature vector $x$. As in the training of ArcFace, we calculate the angle $\theta$ between $x$ and the ground truth weight. We then calculate the logits $\cos(\theta+\theta_m)$ which are scaled by the feature scale $s$ and softmaxed to probabilities. For our membership attack, we calculate the entropy of this vector $p$ for each image by the sum $- \sum_i p_i \log(p_i)$. The entropy distributions are shown in Figure~\ref{fig:aux} and for each method the images are ranked by their calculated entropy. We find that with the ArcFace model trained on MS-Celeb-1M, 78\% of the lowest entropy images belong to MS-Celeb-1M, however with the model trained using our synthetic data set this falls to 51\%. Although this is only a simple attack, it shows that the classifier trained directly on real data exposes the data more clearly than the one trained on generated samples.

\begin{figure}
\centering
\includegraphics[width=1.\textwidth]{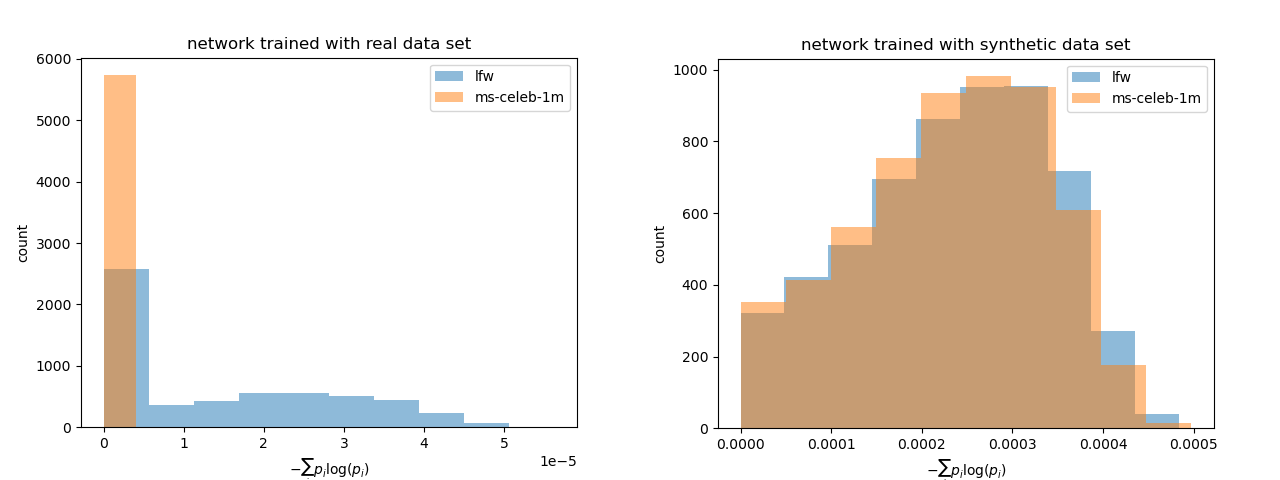}
\caption{Distribution of entropies from data sets LFW and MS-Celeb-1M calculated using networks trained with real and synthetic data.}
\label{fig:membership}
\end{figure}

\subsection{Biases introduced by the synthetic data set}
\newcommand{\genff}{\mathcal{D}_{g}}
\newcommand{\genms}{\mathcal{D}_{gms}}
\newcommand{\ms}{\mathcal{D}_{MS}}
\newcommand{\ffhq}{\mathcal{D}_{FF}}
\newcommand{\lfw}{\mathcal{D}_{LFW}}
In this part, we will try to analyse possible explaining factors for the performance discrepancy between our synthetic data set training of ArcFace and the real data counterpart. To simplify the notation we will note $\genff$ our synthetic data set,
$\ms$ the MS-Celeb-1M data set, $\ffhq$ the FFHQ data set, $\lfw$ the LFW data set. 
The bird's-eye explanation is merely that $\genff$ introduces a larger domain shift with respect to the target validation set $\lfw$ than $\ms$ does.
As a result, ArcFace features are not as robust and informative on $\lfw$ identities when trained on $\genff$ than when trained on $\ms$.

This increased domain shift can be imputed either to 
i) a bias of $\genff$  towards $\ffhq$ due to the GAN loss (which involves $\ffhq$ directly)
or ii) intrinsic biases of the StyleGAN2 network.
To evaluate the latter possibility, we also consider a synthetic data set (denoted by $\genms$) produced by StyleGAN2 trained in a classical way but on $\ms$ instead of $\ffhq$.

\begin{table}[h!]
    \centering
\begin{tabular}{ |c|c|c|c|c|c| } 
 \hline
  & $\genms$ & $\genff$ & $\ms$ & $\ffhq$ & $\lfw$ \\ 
 \hline
 $\genms$ & X & 104.65 & 19.84 & 108.29 & 30.2 \\ 
 \hline
 $\genff$ & 13.42 & X & 85.10 & 11.51 & 110.24\\
 \hline
 $\ms$ & 5.79 & 10.58 & X & 78.85 & 25.01 \\
 \hline
 $\ffhq$ & 6.32 & 9.15 & 0.86 & X & 105.58 \\
 \hline
 $\lfw$ & 6.77 & 11.81 & 1.35 & 2.28 & X\\
\hline 
    \end{tabular}
\caption{Distances between pairs of data sets given in terms of FID (upper half) and Frechet distance in original ArcFace feature space (lower half).}
\label{tab:frechet}
\end{table}

To evaluate domain shifts, we shall rely on the Frechet distance in two feature spaces.
Indeed we report the Frechet distance between pairs of domains in the Inception space yielding the FID and in the ArcFace feature space (trained on $\ms$). 
Since distance matrices are symmetric, we gather the report of both distances in the single table~\ref{tab:frechet}. 
Focusing first on FID (upper half of the table) and in particular the last column, one observes that $\lfw$ is closer to $\ms$ than to  $\ffhq$, which endorses the fact that the domain shift $\lfw\leftrightarrow \ffhq$ is larger than $\lfw\leftrightarrow\ms$.
Note that the margin in terms of FID is severe. 
The same observation holds for the comparison between the domain shifts $\lfw\leftrightarrow\genff$ and $\lfw\leftrightarrow\genms$ with again a large discrepancy. 
These observations validate our hypothesis i).
On the other hand, to put in balance hypothesis i) and ii), one can only note a slight discrepancy of FIDs corresponding to $\lfw\leftrightarrow\ms$ and $\lfw\leftrightarrow\genms$. 
This supports the idea that indeed the SyleGAN2 has an impact in terms of domain shift, but that it is minor compared to the impact of relying on $\ffhq$.

Looking at the distance in the ArcFace feature space (lower half) is also enlightening because those features are mostly equivariant with respect to identity factors. 
In that space, similar observations can be made as in FID by focusing on the last row (again using $\lfw$).
Indeed, whether we consider real data sets or synthetic ones, the reliance on $\ffhq$ increases the distance to $\lfw$.
However in this feature space, the discrepancy is less severe.
This fact indicates that the domain shift introduced by the reliance to $\ffhq$ concerns mostly other factors of variation than those tied to identity. 
This is also confirmed by noting that the ArcFace Frechet distance corresponding to $\ffhq\leftrightarrow\ms$ ($0.86$) is smaller than $\lfw\leftrightarrow\ms$ ($1.35$) while the comparison was reversed in the more generic space of inception features (FID of $78.85$ for $\ffhq\leftrightarrow\ms$ vs $25.01$ for $\lfw\leftrightarrow\ms$).

The outcome of this analysis is that the decrease of performance in the downstream task is rather accounted by intrinsic biases of the subsidiary data set $\ffhq$ rather than by the reliance of StyleGAN2 to produce the synthetic data set.
What is more, these biases concern factors of variations that are uncorrelated with identity, which is the data that we want to protect.
This suggests that gathering a subsidiary data set with less restricted variations could help fill the gap between our current performance and the one obtained by training directly on the private data set $\ms$.

\subsection{Conclusion}

We have proposed an algorithm for creating synthetic datasets for the use of training face verification algorithms. It uses changes in the architecture of the popular StyleGAN2 to enforce the separation of identity and other factors.
We have shown that the GAN generation process does offer an observable improvement in privacy as demonstrated by a simple membership attack. 
A generated dataset which is successful at training a face recognition algorithm also presents the opportunity to introduce a differentially private mechanism into the generation process. This avenue is left for future work, as differential privacy in GANs is not yet mature enough.

Compared to real datasets, our synthetic dataset gives reduced performance on face recognition tasks, with an accuracy of 83\% compared to the state-of-the-art 99.7\%. To investigate the discrepancy in performance we have analysed our synthetic dataset, comparing it to real datasets across a variety of metrics.
We identity a few key issues.
Firstly, the usage of FFHQ in training our GAN could present a problem as the dataset is not perceptually similar to other face dataset.
In particular, FFHQ is quite restricted in terms of non-identity related factors compared to "in the wild" type of datasets. 
As a result, this lack of intra-class diversity affects our synthetic dataset, which is a key problem to tackle in the future.
Finally, the generated dataset may introduce difficulties in the downstream task training, due to the inclusion of generated artifacts which are not perceptually visible but which may still introduce some domain shift.
Our analysis suggests that the latter issue has a lower impact than the former.

We hope that this work provides some interest in tackling the problem of privacy in face recognition datasets. As the size and quality of face datasets increases so does the public's interest in their personal privacy. 
Given our experiments, we endorse that synthetic face datasets provide a promising solution.

\section*{Acknowledgements} This project was partly funded by the support of the ANR through the BIOQOP contract referenced ANR-17-CE39-0006.
\bibliography{egbib}
\end{document}